\documentclass{trbunofficial} % for arXiv

\begin{document}

% ---------- Title Page ----------
% Place Title Page AFTER \begin{document} and BEFORE \maketitle 
% to include the front matter in word count

% Paper title
\title{Calibration-Free Vehicle Speed Estimation: A Monocular Keypoint-Template Approach}

% Add author(s) with: \TRBauthor[*]{Name}{Affiliation}{Email}[Address][ORCID]; Address and ORCID are optional
\TRBauthor{Gaofeng Su}{Department of Civil \& Environmental Engineering\\The University of California, Berkeley}{koufongso@berkeley.edu}
\TRBauthor{Keya Li}{Department of Civil, Architectural and Environmental Engineering\\The University of Texas, Austin}{keya\_li@utexas.edu}
\TRBauthor{Raja Sengupta, Ph.D.}{Department of Civil \& Environmental Engineering\\ The University of California, Berkeley}{rajasengupta@berkeley.edu}
\TRBauthor*{Kara M. Kockelman, Ph.D., P.E}{Department of Civil, Architectural and Environmental Engineering \\ The University of Texas, Austin}{kkockelm@mail.utexas.edu}[512-471-0210]

% Short author line for the running header
\AuthorHeaders{Su, Li, Sengupta, and Kockelman}

\maketitle

\section{Abstract}
This paper proposes a calibration-free framework for reliably and effectively estimating vehicle speeds from monocular videos, without relying on roadway features, camera calibration, or roadway-feature-based reference objects. The proposed framework estimates vehicle speeds using a 36‑keypoint vehicle template and a homography matrix updated at each frame. A YOLO‑based keypoint detection module is trained on diverse datasets, and two estimation strategies are compared: keypoint‑only tracking and warped optical flow with dense spatial aggregation. Speed is referred by projecting displacements into metric space using the homography, with validation conducted on over 400 video clips from road-side and over-head datasets, covering speeds from 30 to 100 mph. The method achieves reliable speed estimation on the V13 and BrnoCompSpeed datasets, with the warped optical flow method delivering MAEs of 15.0\% and 9.7\%, respectively, and 77.9\% and 93.1\% of estimates within ±20\% error. After applying a 10\% trim to remove edge-of-frame outliers, performance improves to 11.7\% and 7.6\% MAE, with within-±20\% accuracy rising to 85.3\% and 95.4\%. This work addresses key limitations of existing vision-based approaches and enables low-cost and efficient speed enforcement using portable devices such as dashcams and smartphones, thereby supporting citizen-based enforcement programs for traffic safety.

\newpage

\newpage

% ---------- Introduction ----------
\section{Introduction}\label{sec:intro}
Estimating vehicle speed accurately is a fundamental topic in Intelligent Transportation Systems (ITS) and plays a critical role in traffic enforcement and improving road safety. Active and automated enforcement of speed limits can lower speeds, crash counts, and injuries \cite{sadeghi2016speed, shin2009evaluation}. For example, fixed speed cameras can reduce 47\% of fatal and injury crashes on urban principal arterials \cite{shin2009evaluation}.

Common speed measurement methods include intrusive technologies (e.g., inductive loop detectors) and non-intrusive technologies (e.g., overhead-mounted and side-fired sensors) \cite{sangsuwan2024video,odat2017vehicle,martin2003detector}. Intrusive detectors are installed within or across the roadway; for example, inductive loop detectors estimate vehicle speeds by detecting vehicle presence within the loop area through changes in oscillator frequency \cite{martin2003detector}. Such technologies are expensive to install and can cause disruptions to traffic flow. For instance, a loop system with 40-cm ducts and three chambers costs approximately \$9{,}000 per approach \cite{cttech_2020_detection}. Non-intrusive technologies include active detectors (such as active infrared, microwave radar, and ultrasonic sensors) and passive detectors (such as passive infrared, acoustic, and video image processing systems). Active detectors transmit energy (e.g., radar waves or LiDAR) toward vehicles and calculate speeds based on the returned signal, while passive detectors measure energy emitted by vehicles \cite{javadi2019vehicle}. Active sensors generally provide longer detection ranges and greater robustness, but require external power sources and are more complex. In contrast, passive sensors are simpler and require lower power consumption but are more susceptible to environmental conditions \cite{hovhannisyan2021_activepassive}. 

Recently, vision-based speed estimation methods have emerged as a more cost-effective and scalable alternative \cite{martinez2024digital}. Over the past decade, computer vision (CV) technologies have been widely used for vehicle detection and tracking \cite{zhang2022real,zuraimi2021vehicle,maity2021faster}, license plate recognition \cite{xu2018towards,montazzolli2017real,parsa2025video}, and speed estimation \cite{hua2018vehicle,liu2020vision}. Most state-of-the-art CV-based speed estimation methods typically rely on camera videos and scaling factors derived from road features to estimate travel distances in a two-dimensional (2D) domain using vanishing points (VP) and bounding boxes \cite{lu20172,kim2018reliability,chang2018deepvp,li2025smartphone}. Despite their effectiveness, these approaches often require prior knowledge or strong assumptions about roadway geometry, fixed roadway features, or camera calibration, which limits their scalability and applicability.

This work introduces a novel and calibration-free framework for vehicle speed estimation that does not rely on roadway features. The proposed approach iteratively updates the homography matrix derived from at least four non-collinear points in the planar video frame, to estimate vehicle speeds and can be readily applied to both overhead- and side-view camera videos. This framework enables speed estimation under flexible deployment conditions and is adaptable to a wide range of roadway environments and camera configurations.

The remainder of this paper is organized as follows. Section II reviews related work on CV-based speed estimation. Section III describes the proposed methodology for vehicle speed estimation. Section IV presents the experimental results on overhead and side‑view public datasets. Finally, Section V concludes the paper and discusses future work.

\section{Related Work}
CV techniques to estimate vehicle speeds typically begin with video recordings and scene parameters (such as image-to-world scaling factors) as inputs, then use detection and tracking algorithms to estimate traveled distances in the 2D image domain. Speeds are calculated using real-world distances derived from scale factors and the time intervals between consecutive frames \cite{gunawan2019detection, abdel2015direct}. One critical component of vision-based speed estimation is camera calibration \cite{zhang2021camera, weng1992camera}, which enables accurate mapping of 2D image coordinates to 3D real-world motion. Calibration involves estimating both intrinsic parameters (sensor size, resolution, focal length) and extrinsic parameters (camera position and orientation). Traditional methods rely on reference objects with known dimensions, such as lane markings or width, but these require manual measurement and are often impractical for large-scale deployment. 

To address this, VP-based calibration has emerged as a widely used approach. VPs occur where parallel lines in the 3D world (e.g., lane boundaries, building edges) appear to converge in the image. VP-based methods are classified into geometry-based and learning-based methods. Geometry-based methods use line intersections, clustering techniques, or Gaussian sphere representations to estimate VPs \cite{feng2010semi, barinova2010geometric, collins1990vanishing}. For instance, \citet{song2021ptz} adopted an improved Direct Linear Transform (DLT) method by obtaining two VPs from vehicle trajectory lines and car body edge lines, combined with a third VP derived from geometric relationships, to calculate speeds. Results show that the error of the proposed method under different conditions was less than 2.5\%. Learning-based methods, instead, use deep neural networks trained on large-scale datasets with VP annotations to infer VPs directly from image context \cite{zhai2016detecting, chang2018deepvp}. \citet{hua2018vehicle} combined deep learning-based object detection with optical flow tracking on the NVIDIA AI City Challenge dataset, which contains over 20 hours of traffic videos from multiple camera angles. Their best results achieved a detection rate score of 1.0 and a root mean square error (RMSE) of 12.109 mph. They also pointed out that performance was particularly poor in locations where camera movement due to wind or bridge vibrations degraded video quality. \citet{rani2024lv} proposed a deep learning‑based framework combining U‑Net segmentation with an LV‑YOLO network to detect logistic vehicles and estimated their speeds from highway CCTV footage. Results indicate that the method achieved a 2.63\% improvement in speed prediction accuracy over the 1D‑CNN speed estimation model.

Based on estimated VPs and standard camera placement assumptions (zero skew, overhead), intrinsic and extrinsic parameters can be derived, enabling transformation between camera and real-world coordinates. Notably, most existing methods are applied to fixed traffic cameras and standard roadway features, which makes speed estimation particularly challenging in environments where roadway features is unclear or unavailable.

\section{Proposed Methodology}
This section presents the methodology for estimating vehicle speeds from monocular video, based on semantic keypoints, optical flow, and adaptive homography-based mapping. The following subsections outline the problem statement, describe the dynamic homography estimation procedure, derive the speed estimation formulation, analyze errors, and present the complete system pipeline.

\subsection{Problem Statement}
Given two consecutive images $I_{k}$ and $I_{k+1}$ capturing $N$ moving vehicles, the objective is to estimate the set of linear speeds $\mathcal{V}_k = \{v_{i,k}\}_{i=1}^N$, where $v_{i,k}$ denotes the instantaneous speed for vehicle instance $i$ at time step $k$. A fundamental challenge in vision-based speed estimation arises from perspective effects, whereby identical pixel displacements do not correspond to equal physical distances in the scene. To resolve this, we utilized homography, a projective transformation that maps points between two planar coordinate frames \cite{hartley2003multiple}. A key advantage of this technique is that it does not require prior camera calibration, provided that at least four non-collinear correspondences between the coordinate frames are identified \cite{song2021ptz}.

\begin{figure}[!htbp]
    \centering
    \includegraphics[width=0.5\linewidth]{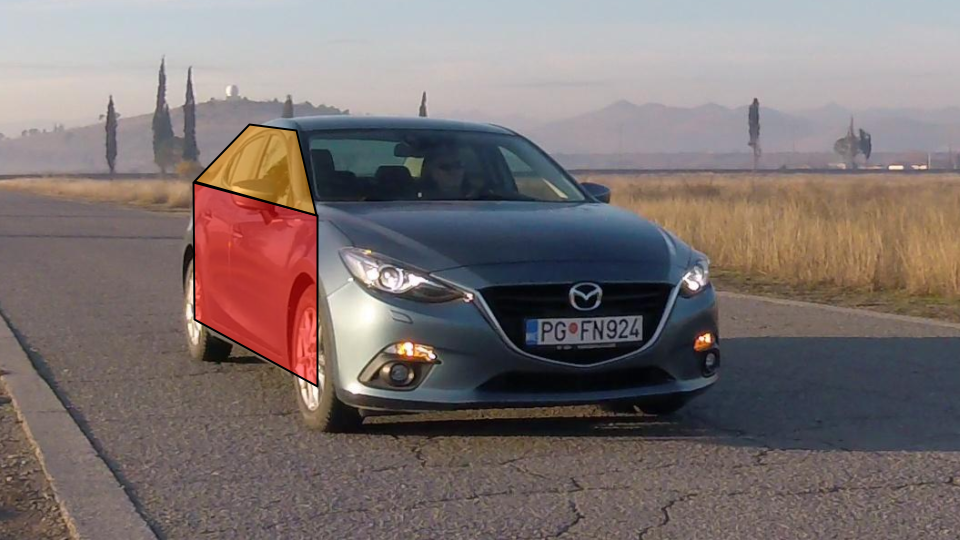}
    \caption{Piecewise planar approximation of a vehicle body. (Frame sampled and modified from the VS13 dataset \cite{djukanovic2022dataset}).}
    \label{fig:vehicle_planar_model}
\end{figure}

Vehicles can be approximated as rigid bodies composed of locally planar surfaces, as illustrated in Figure \ref{fig:vehicle_planar_model}. The red and yellow shaded regions highlight the distinct lateral and upper planar faces modeled, respectively, with an independent coordinate frame established for each plane. Furthermore, it is assumed that the vehicle undergoes longitudinal motion only, with negligible lateral displacement and rotation between consecutive frames. Under this assumption, the motion of each planar surface can be represented by a 2D homography.

\subsection{Notation and Preliminaries}
Let $i \in \{1, \dots, N\}$ denote a specific tracked vehicle instance and $k$ denote the discrete time step. Each vehicle is modeled as a collection of planar facets $c \in \mathcal{C}$, where $\mathcal{C} = \{\text{top, left, right}\}$. Although this work is targeting a multi-vehicle environment, their speeds can be estimated independently through an identical process. Moreover, since the homography transformation applies independently to facets, the subsequent derivation focuses on a single lateral plane of a single vehicle without loss of generality and maintain a clean and understandable notation.

For a tracked vehicle $i$ at time step $k$, we define:
\begin{enumerate}[label=\arabic*)]
\item $I_k$: The image coordinate frame at time $k$. 
\item $V_{k}$: The metric planar coordinate frame attached to a vehicle facet, and is established based on the vehicle's pose at time $k$.
\item $\mathbf{p}_{k} \in I_k$: The pixel coordinates $[u, v]^\top$ of a physical point captured at time $k$.
\item $\mathbf{s}_{k}^{V_{k}}$: The metric coordinates $[x, y]^\top$ of a point captured at time $k$, expressed in the metric frame $V_{k}$.
\item $\mathcal{M}_{k}$: The set of detected semantic keypoints belonging to a facet at time $k$.
\item $\mathbf{F}_k$: The dense optical flow field, where $\mathbf{p}_{k+1} = \mathbf{p}_{k} + \mathbf{F}_k(\mathbf{p}_{k})$ represents the correspondence of the same physical point across consecutive image frames.
\item $v_{k}$: The final estimated scalar speed of the vehicle at time $k$.
\end{enumerate}

\subsection{Dynamic Image-Template Homography Estimation}

The projective relationship between frame $I_k$ and $V_k$ is modeled by a homography matrix $\mathbf{H}_{k} \in \mathbb{R}^{3 \times 3}$. Using homogeneous coordinates, the mapping from image space to the metric vehicle plane is given by:

\vspace{-12pt}
\begin{equation}
\label{eq:homography_projection}
w \begin{bmatrix} x \\ y \\ 1 \end{bmatrix} = \mathbf{H}_{k} \begin{bmatrix} u \\ v \\ 1 \end{bmatrix}
\end{equation}

or equivalently,
\vspace{-2pt}
\begin{align}
\label{eq:homography_projection_compact}
& w \mathbf{\tilde{s}}_{k}^{V_{k}} = \mathbf{H}_{k} \tilde{\mathbf{p}}_{k}
\end{align}

where $w \neq 0$ is a projective scale factor. The corresponding Euclidean coordinates on the vehicle plane are recovered by normalizing the homogeneous representation with respect to the third component, and we expressed it as:

\vspace{-12pt}
\begin{align}
\label{eq:homography_projection_normalization}
& \mathbf{s}_{k}^{V_{k}} = \sigma(\mathbf{H}_{k} \tilde{\mathbf{p}}_{k})
\end{align}

The homography matrix $\mathbf{H}_{k}$ can be estimated from at least four non-collinear correspondences between the semantics keypoint $p_k\in \mathcal{M}_{k}$, and their corresponding locations on a physical vehicle metric template. When more than four correspondences are available, $\mathbf{H}_{k}$ is estimated by solving an overdetermined system, with Random Sample Consensus (RANSAC) algorithm to reject outliners \cite{fischler1981random}. The mathematical formulation of homography estimation is well established and is therefore omitted here \cite{hartley2003multiple}.

We defined a vehicle metric template represented by a 2D mechanical drawing labeled with 36 semantic keypoints. Each keypoint is associated with 2D coordinates in meters. Figure \ref{fig:keypoints} illustrates the sedan-based metric template used in our experiments. To preserve the planar assumption underlying Equation ~\eqref{eq:homography_projection}, we selected subsets of approximately co-planar keypoints corresponding to a specific vehicle facet, for example, wheels center, doors, door handle, side window corners. By associating detected image keypoints $\mathbf{p}_{k}$ with their corresponding metric template coordinates $\mathbf{s}_{k}^{V_k}$, an instantaneous homography matrix  $\mathbf{H}_{k}$ for each frame $k$ was estimated. Because the metric scale is enforced by known dimensions of the canonical template,  $\mathbf{H}_{k}$ provides an image-to-metric mapping that adapts to the vehicle's current position within the field of view.

While homography-based methods can theoretically accommodate viewpoint changes, their accuracy relies on the assumption that the observed motion on a vehicle surface can be represented by a single rigid plane. In practice, vehicle surfaces are only approximately planar, and vehicle motion introduces deviations due to suspension dynamics, minor rotations, and depth variations across keypoints. These effects cause the true image-to-plane transformation to gradually deviate from the estimated homography, resulting in persistent metric errors. Moreover, the homography is estimated from keypoint correspondences at a specific vehicle pose and is therefore only valid for that reference configuration. As the vehicle moves, the tracked keypoints may undergo increasingly large image displacements relative to their original positions, while the fixed homography cannot account for the resulting changes in projection. Consequently, planar approximation errors and temporal homography mismatch accumulate over time, leading to drift in displacement and speed estimates.

To address this limitation, this paper models the homography as a dynamic, time-varying transformation. By re-estimating the homography at each time step using keypoints associated with the same vehicle planar facet, the proposed approach continuously adapts the planar approximation to the vehicle’s current pose. This dynamic estimation reduces the accumulation of projection bias and improves the consistency of displacement and speed estimates over time.

\subsection{Speed Estimation via Warped Optical Flow}

At time step $k$, a planar metric coordinate frame $V_k$ is defined for each
tracked vehicle facet. Given a homography $\mathbf{H}_k$ between the image
frame $I_k$ and the metric planar frame $V_k$, the metric displacement of an
arbitrary point on the facet can be recovered if its image correspondence
between consecutive frames is known. Specifically, for a point
$\mathbf{p}_{i,k}$ and its corresponding location $\mathbf{p}_{i,k+1}$ at the
next time step, the instantaneous speed is computed as

\vspace{-12pt}
\begin{align}
\label{eq:warp_optical_flow_single}
v_{s_{i,k}} =
\frac{
\left\|
\sigma(\mathbf{H}_{k}\tilde{\mathbf{p}}_{i,k+1})
-
\sigma(\mathbf{H}_{k}\tilde{\mathbf{p}}_{i,k})
\right\|
}
{\Delta t},
\end{align}

where $\tilde{\mathbf{p}}$ represents the point in homogeneous coordinates,
$\sigma(\cdot)$ denotes the homogeneous normalization operator, and
$v_{s_{i,k}}$ represents the metric speed of the tracked point. Note that
$\mathbf{H}_{k}$ is only valid for points belonging to the same planar facet
used during its estimation.

The required point correspondence can be obtained using either dense or
sparse tracking strategies. For dense correspondence, the optical flow field
$\mathbf{F}_k$ is used to obtain the subsequent image location:

\vspace{-12pt}
\begin{equation}
\mathbf{p}_{i,k+1}
=
\mathbf{p}_{i,k}
+
\mathbf{F}_k(\mathbf{p}_{i,k}).
\end{equation}

Alternatively, sparse correspondence can be obtained by detecting and
matching semantic keypoints that appear in both consecutive frames. This
provides direct correspondences between $\mathbf{p}_{i,k}$ and
$\mathbf{p}_{i,k+1}$ without relying on dense optical flow estimation.

For the optical flow strategy, a convex hull $\mathcal{S}_k$ is constructed from
the detected semantic keypoints $\mathbf{p}_{k}\in\mathcal{M}_{k}$ to restrict
the queried points to the same planar facet. Optical flow is then evaluated
only for points within this convex hull, ensuring that the resulting
correspondences remain valid under the facet-specific homography $\mathbf{H}_k$.
Moreover, this constraint increases the number of available tracking points beyond
the sparse semantic keypoints, enabling robust aggregation over multiple
measurements and reducing the influence of individual optical flow outliers.

The final vehicle speed estimate is obtained by aggregating the point-wise
speed estimates over all tracked points in $\mathcal{S}_k$:

\begin{equation}
v_k =
\operatorname{RobustAgg}
\left(
\{v_{s_{i,k}} \mid s_{i,k}\in\mathcal{S}_k\}
\right),
\end{equation}

where $\operatorname{RobustAgg}(\cdot)$ denotes a robust aggregation function
used to reduce the influence of optical flow outliers and keypoint localization
errors.

\begin{figure}[!htbp]
    \centering
    \includegraphics[width=0.9\linewidth]{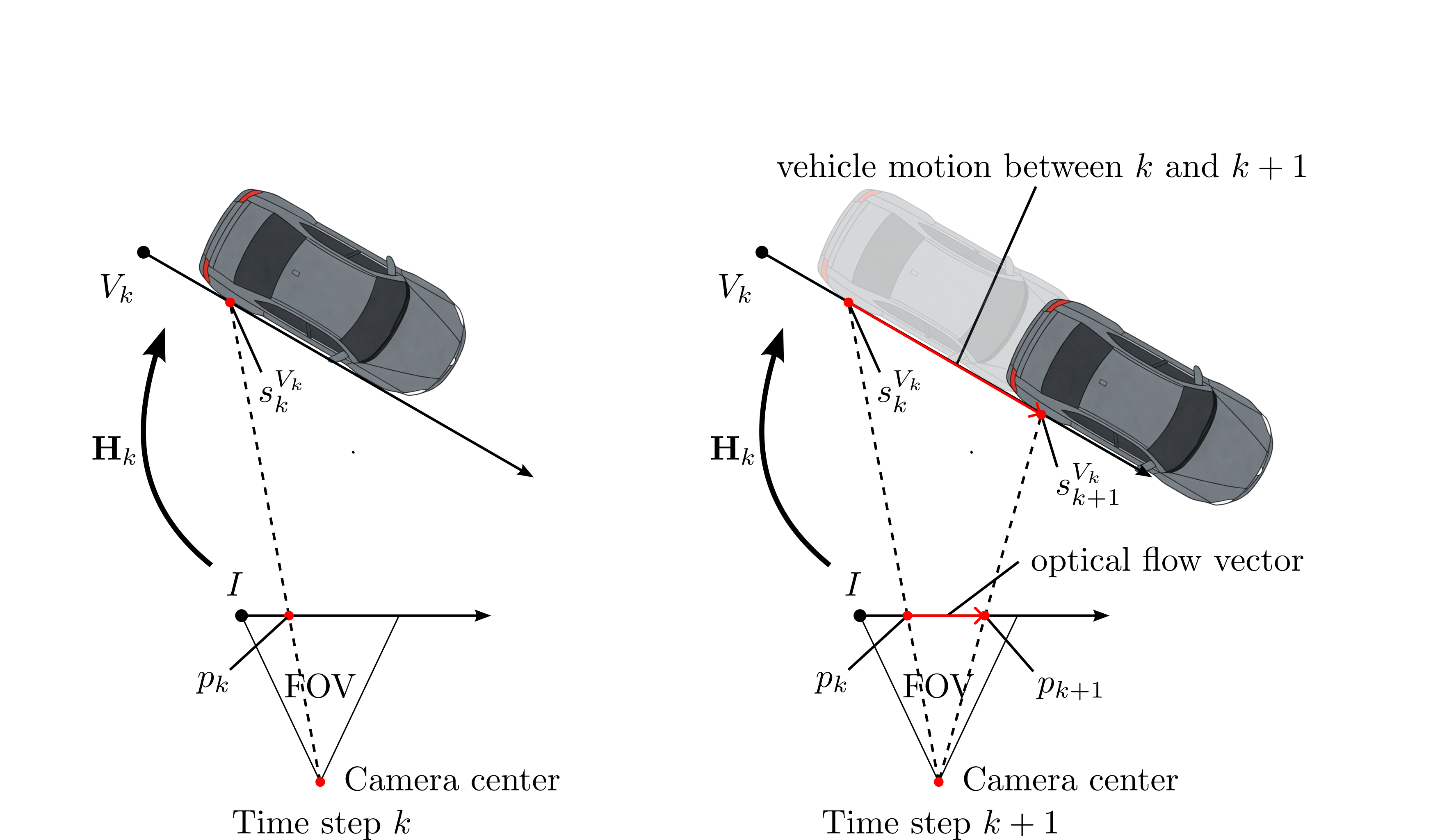}
    \caption{Illustration of speed estimation via warped optical flow for a tracked vehicle's right facet. The vehicle graphic was generated using AI tools.}
    \label{fig:warp_optical_flow}
\end{figure}

\subsection{Error Analysis}
The speed estimate in (\ref{eq:warp_optical_flow_single}) depends on the accuracy of the estimated homography matrix, which is recovered from image keypoint correspondences, and the accuracy of the optical flow. To quantify this effect, we analyzed the first-order sensitivity of the proposed speed estimation method to homography estimation errors.

For notation simplicity, the subscript $k$ is dropped in the following analysis. Let $H^*$ denote the true homography matrix, and $\Delta H$ denote the homography matrix estimation error, yielding:

\vspace{-16pt}
\begin{align}
    H = H^* + \Delta H
\end{align}

Similarly, let $s_1 = s_0 + v_{OF}^*$ represents the true correspondence between two frames, where $v_{OF}^*$ is the true optical flow vector. Accounting for optical flow error $\Delta v_{OF}$, we have:

\vspace{-16pt}
\begin{align}
    v_{OF} = v_{OF}^* + \Delta v_{OF}
\end{align}

Using Equation \eqref{eq:warp_optical_flow_single}, the true velocity $v^*$ satisfies: 
\begin{align}
    v^* \Delta t = \sigma(H^*s_1^*)-\sigma(H^*s_0^*) 
\end{align}

While the estimated velocity $v$ is computed as: 

\vspace{-16pt}
\begin{align}
    v &= \Delta t ^{-1}(\sigma(Hs_1)-\sigma(Hs_0)) \\
      &= \Delta t ^{-1}(\sigma((H^* + \Delta H)s_1)-\sigma((H^* + \Delta H)s_0))
\end{align}

The first-order approximation of the normalization function $\sigma$ gives: 
\begin{align}
    \sigma((H^* + \Delta H)s) \approx \sigma(H^*s) + J_\sigma(H^*s)\Delta Hs
\end{align}

where $J_\sigma$ is the Jacobian of the normalization operator $\sigma$. Consequently, 

\vspace{-14pt}
\begin{align}
    v - v^* \approx \Delta t ^{-1} \big( J_\sigma(H^*s_1)\Delta Hs_1 - J_\sigma(H^*s_0)\Delta Hs_0 \big)
\end{align}

Assuming small inter-frame motion such that $J_\sigma(H^*s_1) \approx J_\sigma(H^*s_0) = J$, it follows that:

\vspace{-14pt}
\begin{align}
    v - v^* \approx \Delta t ^{-1}J\Delta H(s_1-s_0) = \Delta t ^{-1}J\Delta H v_{OF}
\end{align}

The error is bounded by:
\begin{align}
\label{eq:error_bound}
    \lVert\Delta v\rVert = \lVert v - v^*\rVert \leq \Delta t ^{-1} \lVert J\rVert \lVert \Delta H\rVert \lVert v_{OF} \rVert
\end{align}

The bound in (\ref{eq:error_bound}) indicates that the speed estimation error is affected by three components: the perspective sensitivity $\|J_\sigma\|$, the homography estimation error $\|\Delta H\|$, and the magnitude of the optical flow $\|v_{OF}\|$. The first term reflects geometric amplification driven by the viewing angle between the vehicle facet and the image plane. The second term depends on the conditioning of the linear system used to solve for the homography. The third term shows that larger image motions linearly scale the propagation of homography errors into the final metric velocity.

\subsection{Integrated System Pipeline}

\begin{figure}[!htbp]
    \centering
    \includegraphics[width=0.9\linewidth]{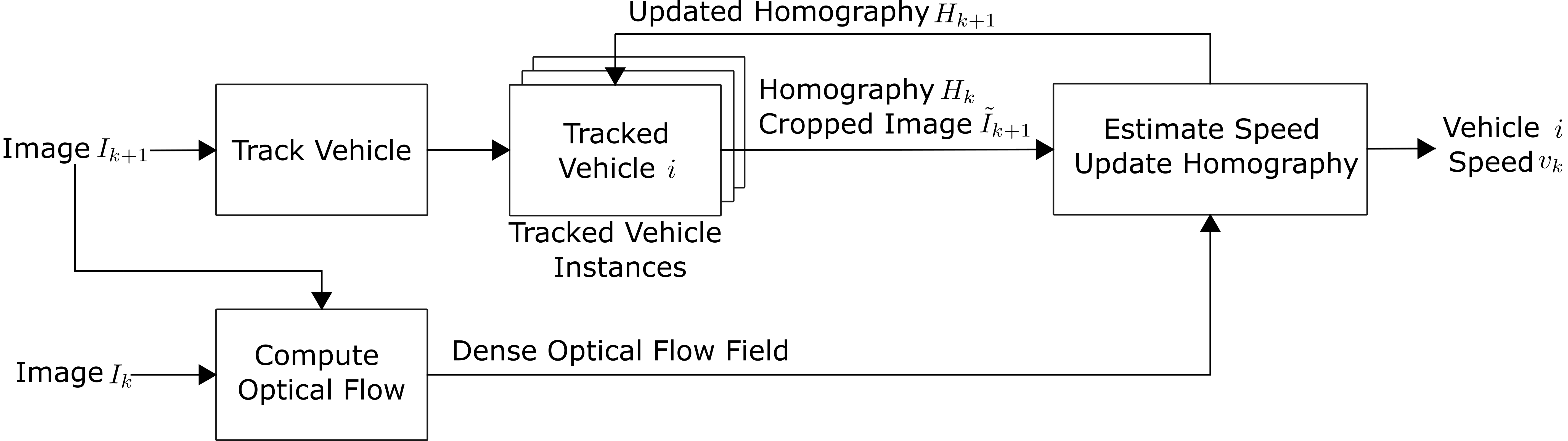}
    \caption{Proposed System Architecture. }
    \label{fig:system_block_diagram}
\end{figure}

Building upon the theoretical framework established in the previous section, the architecture of the proposed integrated system is illustrated in Figure \ref{fig:system_block_diagram}. The pipeline processes a sequence of video frames to simultaneously extract dense motion features and semantic vehicle attributes.

For each incoming frame pair $\{I_k, I_{k+1}\}$, a dense optical flow field $\mathbf{F}_k$ is computed to capture pixel-level displacements. In parallel, a Multi-Object Tracking (MOT) module identifies, localizes and generate bounding boxes for each vehicle. Based on these bounding boxes, the image is cropped to generate vehicle-specific Regions of Interest (ROIs). This spatial isolation ensures that keypoints are detected exclusively within the context of the target vehicle, thereby minimizing background noise and enhancing the accuracy of the detection model.

% shorten the name of the figure: The pipeline processes consecutive frames $I_k$ and $I_{k+1}$ to compute a dense optical flow field. Simultaneously, multiple vehicle instances are tracked to generate cropped ROIs, denoted as $\tilde{I}_{k+1}$. The core Speed Estimation module utilizes the current homography state $H_k$ to warp pixel displacements into metric speeds $v_{i,k}$, subsequently updating the homography to $H_{k+1}$ for the next temporal step.

The core component of the proposed system is the \textit{Estimate Speed and Update Homography} module, which integrates semantic keypoints, optical flow, and adaptive homography estimation for metric speed recovery. At each iteration, the module executes a three-step procedure:

\begin{enumerate}[label=\arabic*)]
    \item \textbf{Keypoint Detection and Classification:} First, a specialized keypoint detection network identifies semantic keypoints within the vehicle ROI and classifies them according to predefined vehicle facets (e.g., the left wheel center keypoint belongs to the left facet).
    
    \item \textbf{Speed Estimation:} For each facet, the system retrieves the associated keypoints and constructs the convex hull $\mathcal{S}_k$, employing the homography matrix $\mathbf{H}_k$ established at the previous time step. Using the inter-frame time interval $\Delta t$ and the optical flow vectors, the metric speed for each point $s_{i,k} \in \mathcal{S}_k$ is calculated via Equation \ref{eq:warp_optical_flow_single}. To determine the final vehicle speed, we aggregate the individual point speeds using an arithmetic mean. Note that more robust aggregation functions (e.g., RANSAC or median filtering) may be employed here to mitigate the influence of outliers.
    
    \item \textbf{Geometric Calibration (Homography Update):} Finally, the homography matrix is updated for the next iteration. For any facet containing a sufficient number of valid keypoints (specifically, $N \ge 4$ non-collinear points), the homography is recomputed to account for changes in the vehicle's perspective or road geometry.
\end{enumerate}

This recursive estimate-then-update strategy ensures that the metric scaling remains tightly coupled to the vehicle's evolving perspective and distance. By continuously refining the homography matrices based on new keypoint detections, the system effectively resets accumulated tracking error for every frame in the sequence, mitigating drift over long tracking durations.

It is important to note that this design introduces a deterministic one-frame latency, as the speed estimate for time step $k$ relies on the optical flow derived from the arrival of the subsequent frame $I_{k+1}$. While technically acausal with respect to the instantaneous frame $I_k$, this forward-looking dependency is inherent to dense optical flow algorithms, which require temporal displacement to compute motion. However, given the high temporal resolution of modern camera hardware, typically $\geq$ 30 frames per second (FPS), this results in a negligible latency of approximately 33 ms. This delay is well within the acceptable operational bounds for real-time applications such as traffic flow analysis and automated speed enforcement.

\section{Experiment Result and Disccusion}

Experiments were conducted to evaluate the effectiveness of the proposed vehicle speed estimation framework. The evaluation focuses on the accuracy of the recovered vehicle speeds, the robustness of the proposed homography-based metric mapping, and the performance of the system under varying vehicle poses and camera viewpoints.

\begin{figure}[!htbp]
    \centering
    \includegraphics[width=0.6\linewidth]{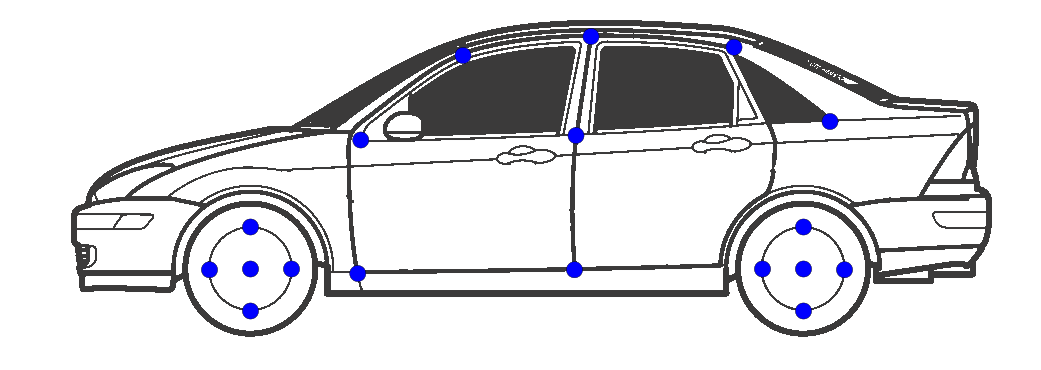}
    \caption{36 semantic keypoints for the sedan are marked with blue circles (symmetric on the other side). Assuming a wheelbase of approximately 3 meters. Car graphics were adapted from Wikipedia. \cite{wiki_car_image}}
    \label{fig:keypoints}
\end{figure}

A sedan model consisting of 36 semantic keypoints was used as the metric template, as shown in Figure \ref{fig:keypoints}. Each keypoint was assigned a 2D (image) coordinates in meters, assuming a wheelbase (horizontal) distance of 3 meters. The keypoint detection module was trained using the Ultralytics YOLO framework with the YOLO-11s backbone \cite{Jocher_Ultralytics_YOLO26_Unified_2026}. The training dataset consists of 756 images for training and 247 images for validation, collected from the Waymo Open Dataset \cite{sun2020scalability}, Roundabout HD dataset \cite{lin2025roundabouthd}, and BoxCar dataset \cite{sochor2018boxcars}.

\begin{figure}[htbp]
    \centering
    \includegraphics[width=0.45\linewidth]{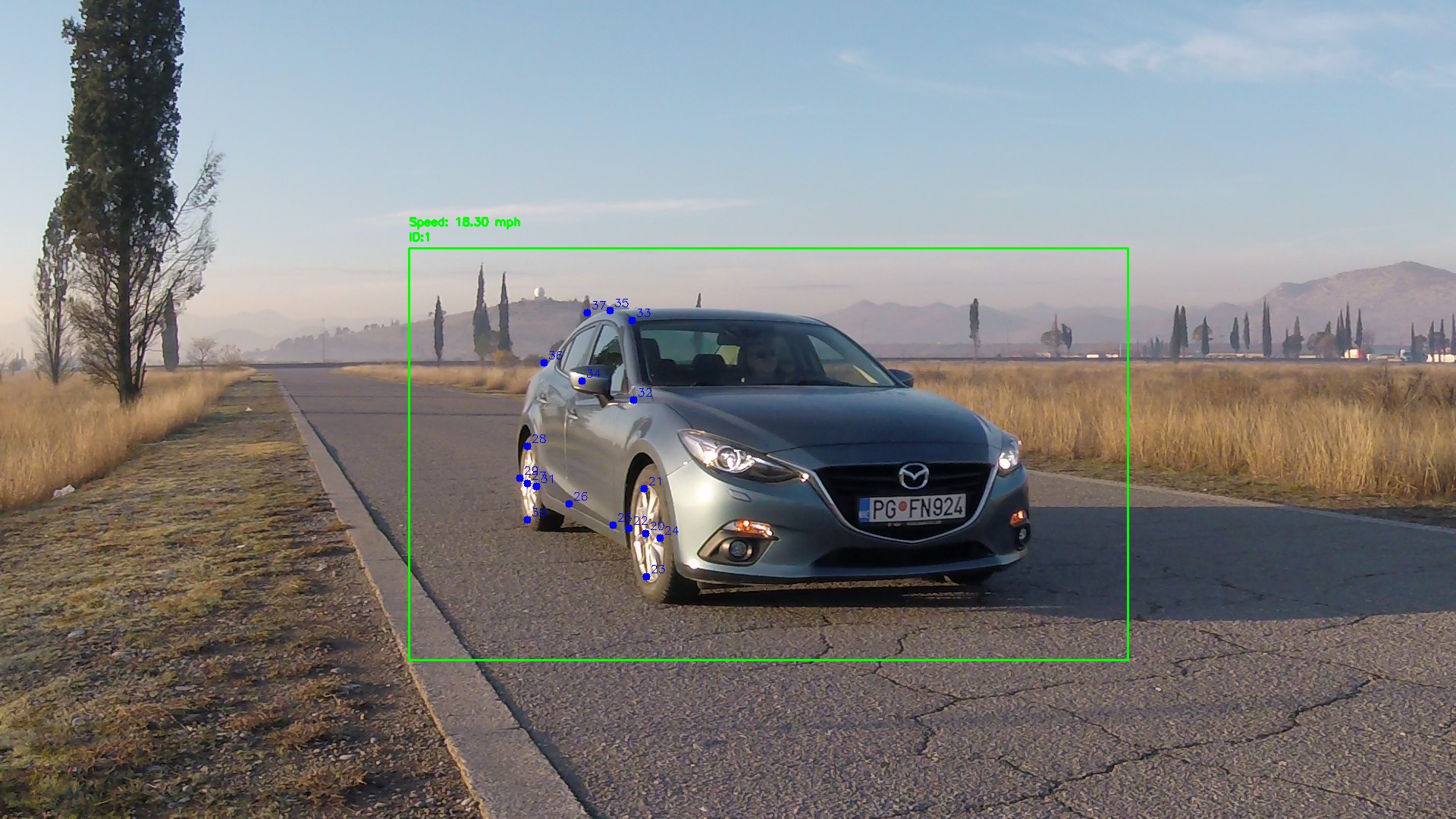}
    \hfill
    \includegraphics[width=0.45\linewidth]{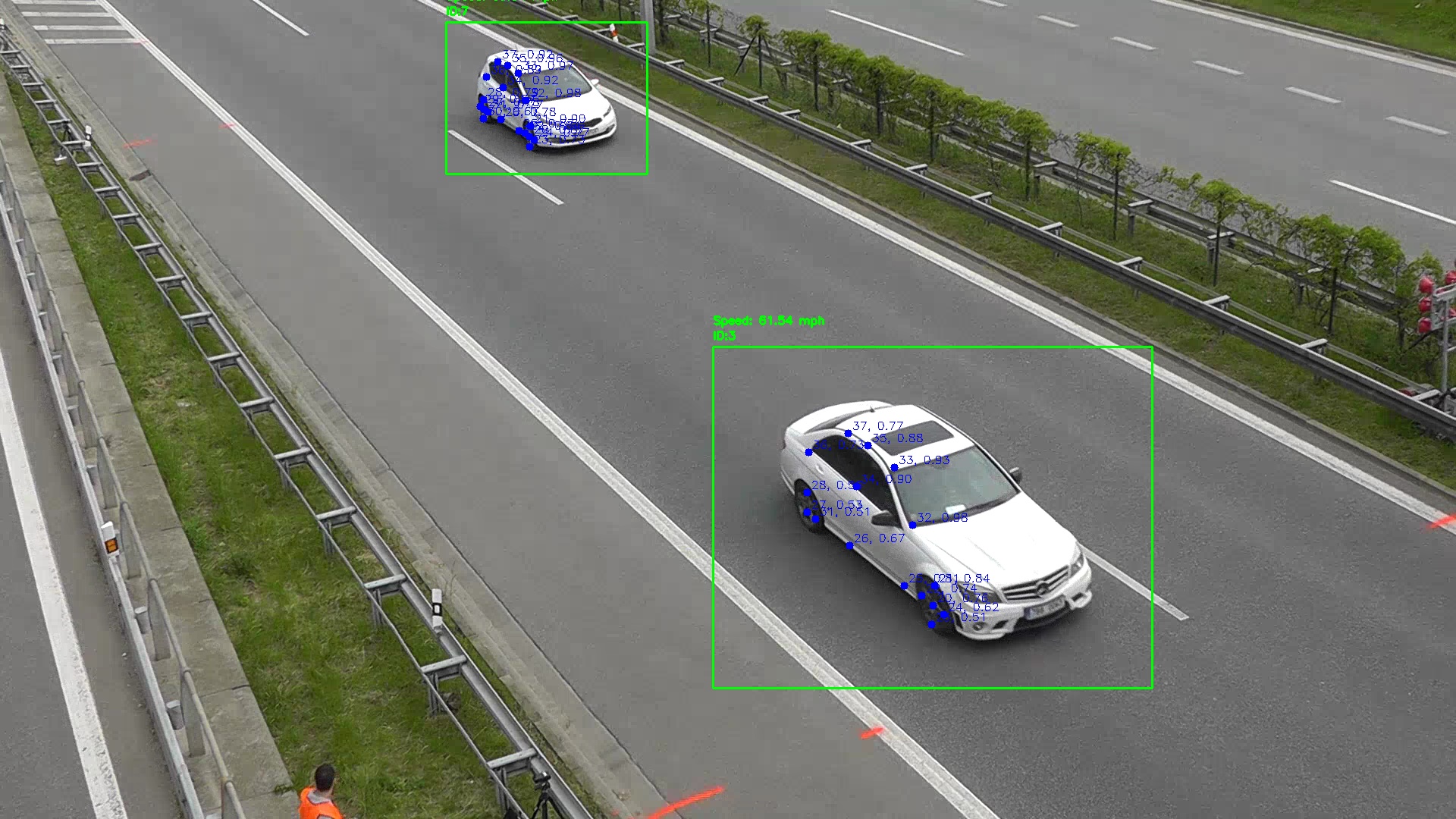}
    \caption{Sample image from the VS13 dataset (left) and BrnoCompSpeed dataset (right), annotated using the speed estimation framework.}
    \label{fig:two_images}
\end{figure}

The validation experiments were conducted using video sequences from the VS13 (resolution 1920 $\times$ 1080, 30FPS) \cite{djukanovic2022dataset} and BrnoCompSpeed datasets (resolution 1920 $\times$ 1080, 50FPS) \cite{sochor2018comprehensive}, with total 443 video clips across 208 unique vehicles and speed ranging from 30 mph to 100 mph. None of these evaluation sequences were part of the keypoint detection model's training data. The two datasets offer two different camera viewpoints: VS13 provides pedestrian-level angles of approaching traffic, while BrnoCompSpeed provides overhead surveillance views. This allows us to test the framework across different viewing geometries and motion profiles. Figure \ref{fig:two_images} shows sample processed frames from each dataset, with tracked cars marked by green bounding boxes and detected semantic keypoints highlighted as blue circles. We also compare two distinct speed estimation strategies: a Keypoint-only (KP) approach that tracks sparse semantic features, and our proposed Warped Optical Flow (OF) method, which leverages a keypoint-defined convex hull to achieve robust, dense spatial aggregation. All errors are presented as relative percentages for comparison across different speeds.

\begin{table}[!htbp]
    \centering
    \small
    \caption{Summary of speed estimation accuracy for VS13 and BrnoCompSpeed datasets.}
    \label{tab:results_combined}
    \begin{tabular}{lllrrrrrrrr}
        \hline
        \multirow{2}{*}{\textbf{Data}} & \multirow{2}{*}{\textbf{Method}} & \multirow{2}{*}{\textbf{Group}} & \multirow{2}{*}{\textbf{N}} & \multirow{2}{*}{\textbf{Med.}} & \multirow{2}{*}{\textbf{MAE}} & \multirow{2}{*}{\textbf{RMSE}} & \multicolumn{4}{c}{\textbf{Thresholds}} \\
        \cline{8-11}
        & & & & & & & \textbf{$\pm$5\%} & \textbf{$\pm$10\%} & \textbf{$\pm$20\%} & \textbf{$>$50\%} \\
        \hline
        \multirow{6}{*}{VS13} 
            & \multirow{3}{*}{KP} 
                & All & 4,317 & -0.9\% & 20.8\% & 178.4\% & 46.7\% & 72.5\% & 87.6\% & 4.4\% \\
                & & Sedan & 2,055 & -0.7\% & 24.3\% & 243.0\% & 46.0\% & 73.3\% & 89.7\% & 3.2\% \\
                & & NoSed & 2,262 & -1.0\% & 17.7\% & 84.2\% & 47.3\% & 71.7\% & 85.7\% & 5.6\% \\
            \cline{2-11}
            & \multirow{3}{*}{OF} 
                & All & 4,578 & -6.0\% & 15.0\% & 53.5\% & 37.3\% & 58.8\% & 77.9\% & 5.3\% \\
                & & Sedan & 2,195 & -6.1\% & 14.3\% & 22.1\% & 35.4\% & 56.8\% & 76.1\% & 4.8\% \\
                & & NoSed & 2,383 & -6.0\% & 15.6\% & 71.1\% & 39.1\% & 60.7\% & 79.6\% & 5.7\% \\
        \hline
        \multirow{6}{*}{BC} 
            & \multirow{3}{*}{KP} 
                & All & 17,085 & 6.5\% & 12.8\% & 46.9\% & 28.3\% & 54.3\% & 87.9\% & 1.2\% \\
                & & Sedan & 3,361 & 6.9\% & 11.1\% & 28.8\% & 32.0\% & 58.9\% & 89.0\% & 0.6\% \\
                & & NoSed & 13,724 & 6.3\% & 13.2\% & 50.4\% & 27.3\% & 53.1\% & 87.6\% & 1.3\% \\
            \cline{2-11}
            & \multirow{3}{*}{OF} 
                & All & 17,600 & 3.0\% & 9.7\% & 66.1\% & 41.4\% & 70.6\% & 93.1\% & 1.0\% \\
                & & Sedan & 3,452 & 3.9\% & 11.0\% & 141.0\% & 47.5\% & 75.1\% & 93.9\% & 0.7\% \\
                & & NoSed & 14,148 & 2.7\% & 9.3\% & 24.2\% & 39.9\% & 69.5\% & 92.9\% & 1.0\% \\
        \hline
    \end{tabular}
    \smallskip
    {\footnotesize \textit{Note: KP = Keypoints; OF = Optical Flow. Med. = Median; MAE = Mean Absolute Error; RMSE = Root Mean Square Error; N = Number of frames with valid estimation.}}
\end{table}

\begin{figure}[!htbp]
    \centering
    \begin{subfigure}[b]{0.47\linewidth}
        \centering
        \includegraphics[width=\linewidth]{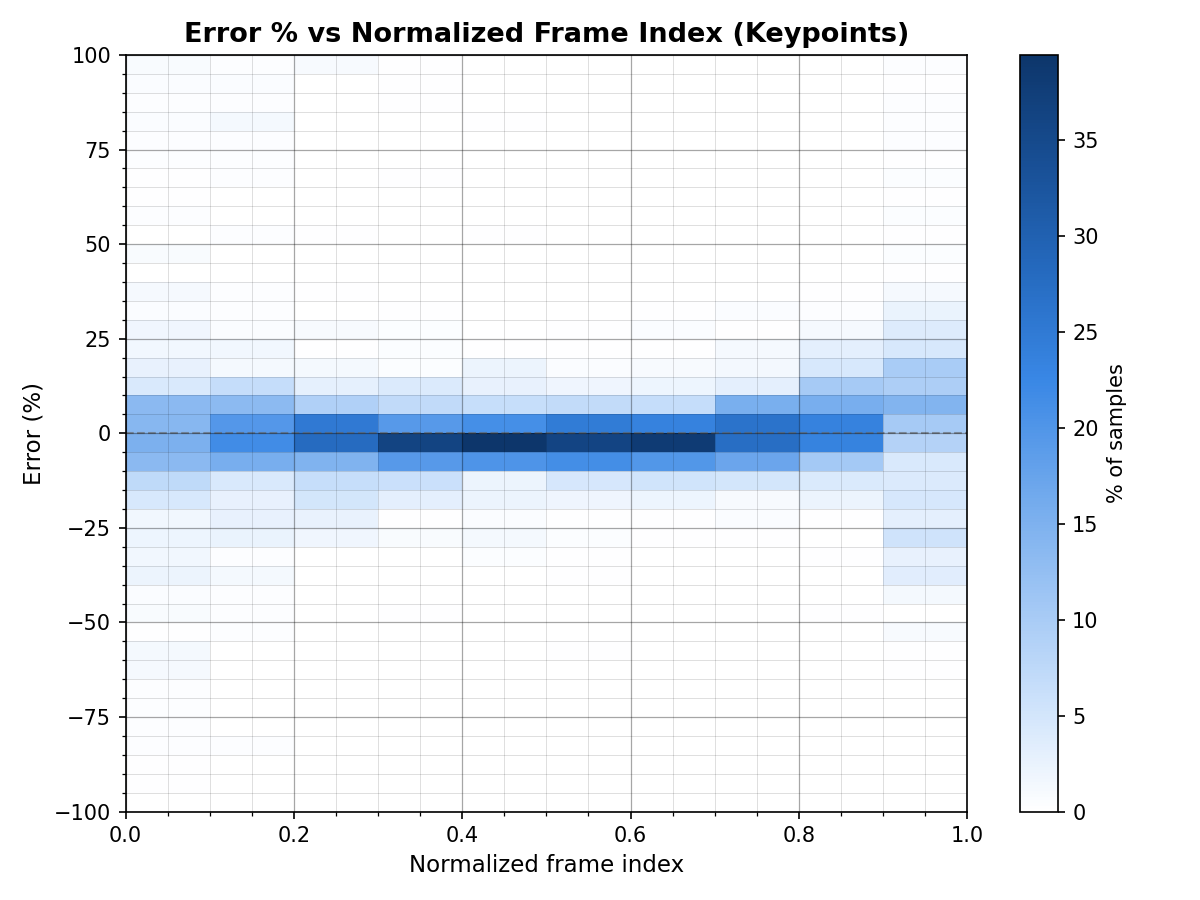}
        \caption{VS13 Keypoints}
        \label{fig:vs13_1}
    \end{subfigure}
    \hfill
    \begin{subfigure}[b]{0.47\linewidth}
        \centering
        \includegraphics[width=\linewidth]{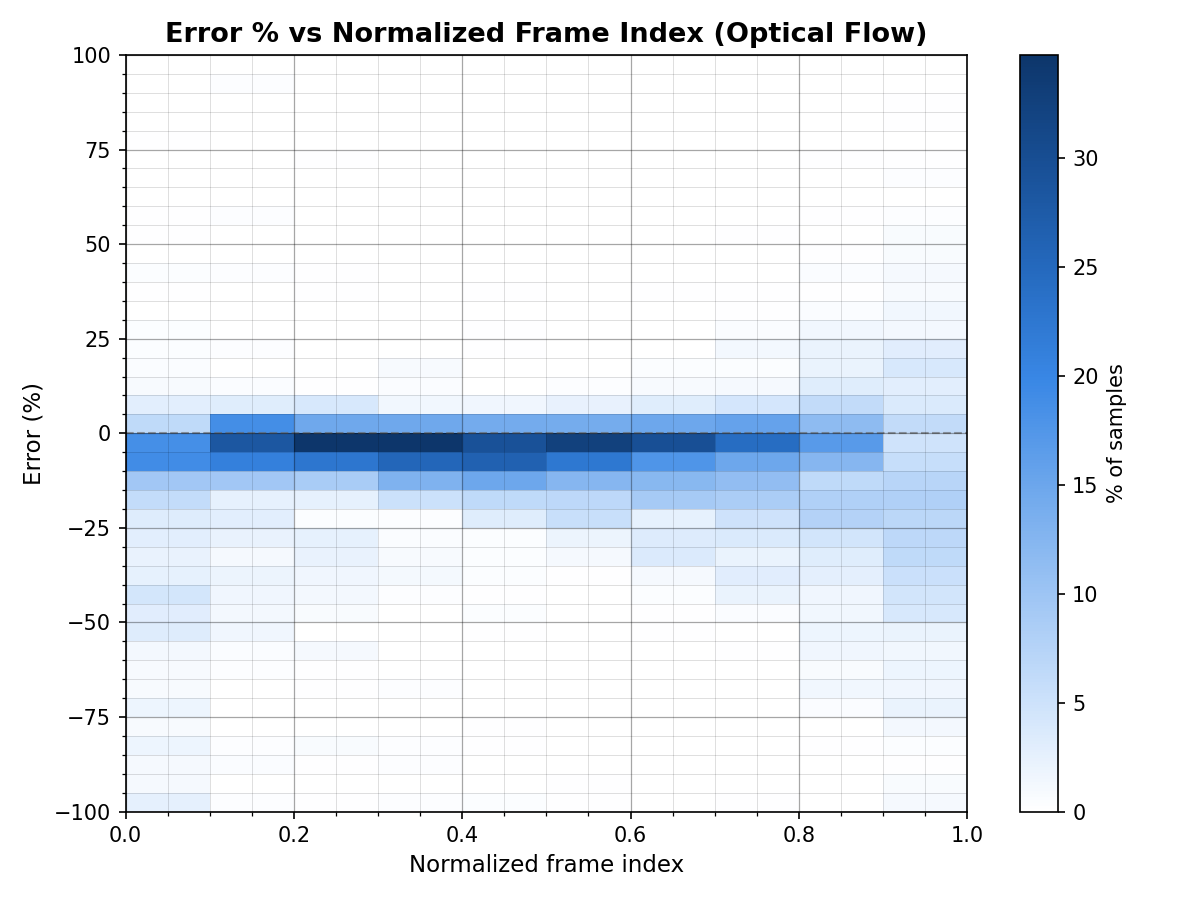}
        \caption{VS13 Optical Flow}
        \label{fig:vs13_2}
    \end{subfigure}
    \\
    \begin{subfigure}[b]{0.47\linewidth}
        \centering
        \includegraphics[width=\linewidth]{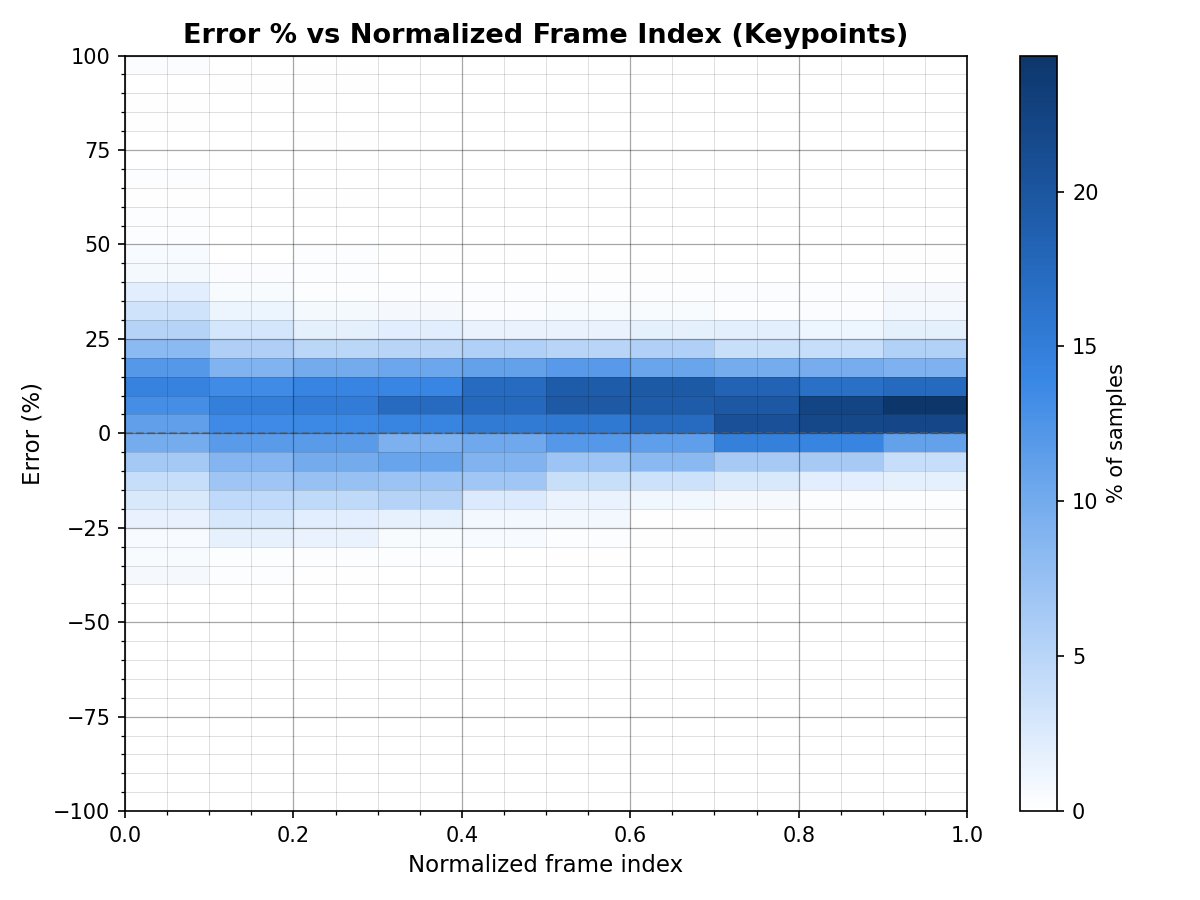}
        \caption{BrnoCompSpeed Keypoints}
        \label{fig:brno_1}
    \end{subfigure}
    \hfill
    \begin{subfigure}[b]{0.47\linewidth}
        \centering
        \includegraphics[width=\linewidth]{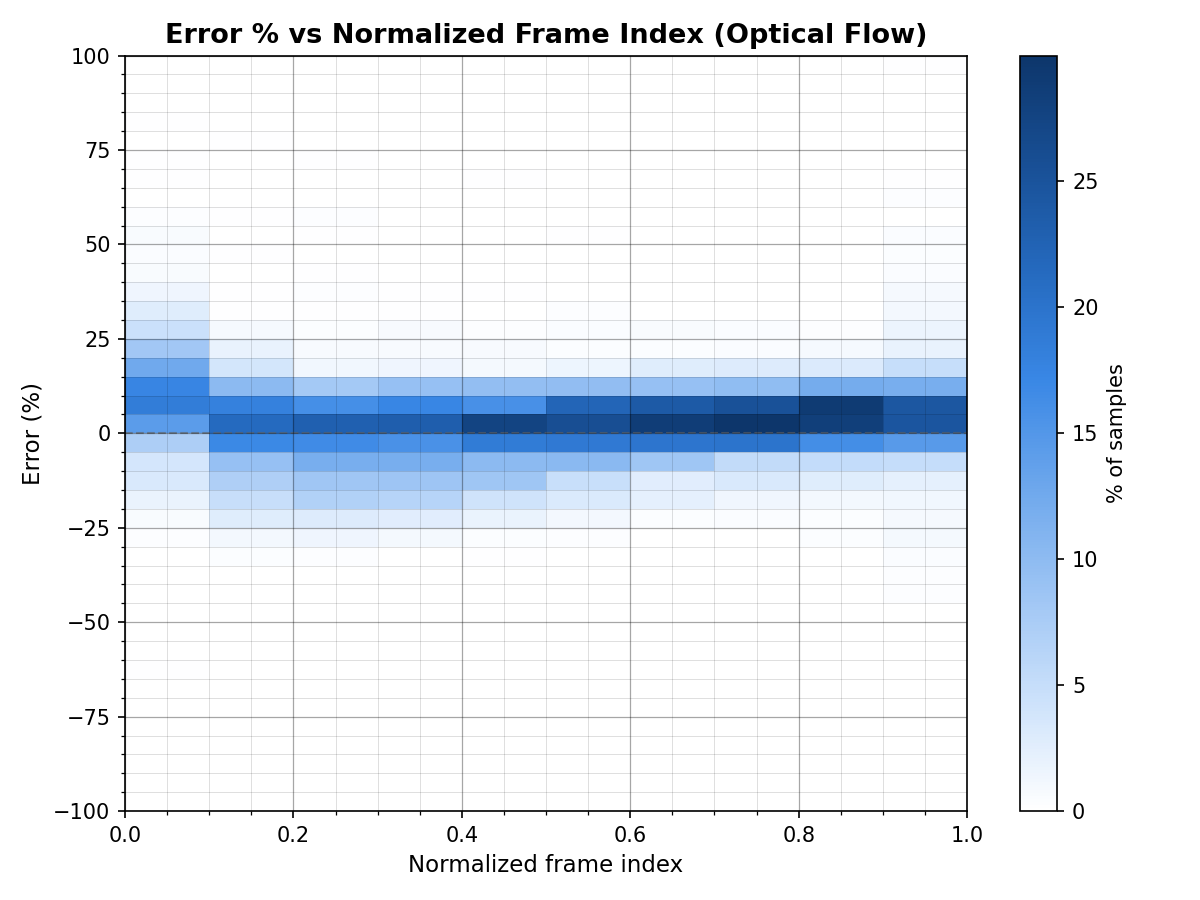}
        \caption{BrnoCompSpeed Optical Flow}
        \label{fig:brno_2}
    \end{subfigure}
    \caption{Error Percentage vs. Frame Index for VS13 and BrnoCompSpeed datasets}
    \label{fig:four_images}
\end{figure}

Table \ref{tab:results_combined} presents the overall accuracy of speed estimation in both datasets. For the VS13 dataset, the optical flow method achieves a Mean Absolute Error (MAE) of 15.0\% with a median error of--6.0\% and 77.9\% of estimates within ±20\% error, compared to the keypoint method which achieves an MAE of 20.8\% with a median error of -0.9\% and 87.6\% within ±20\%. Although the keypoint method achieves higher within-threshold accuracy, which is particularly relevant for speed limit enforcement, the optical flow method yields significantly lower MAE and RMSE, indicating more consistent and accurate estimates overall. For the BrnoCompSpeed dataset, both methods perform significantly better, with optical flow achieving an MAE of 9.7\% and 93.1\% within ±20\%, compared to 12.8\% MAE and 87.9\% within ±20\% for the keypoint method. The optical flow method consistently outperforms the keypoint approach across all metrics on this dataset. Across both datasets, optical flow demonstrates higher within-threshold accuracy. Performance differences between sedans and non-sedans are very small, with sedans generally achieving slightly lower MAE with optical flow, likely due to the dataset being more trained on sedans.

To investigate the temporal behavior of estimation errors, the error percentage was analyzed as a function of the normalized frame index. Specifically, frame indices were scaled to a range of 0 to 1 based on the first and last frames containing a valid speed estimation. Figure \ref{fig:four_images} displays heatmaps of error percentage versus normalized frame index for both methods and datasets.  The heatmaps indicate that errors were significantly higher at the beginning and the end of each video sequence, when vehicles were either too far from the camera—resulting in a small pixel footprint—or too close, partially leaving the frame and making feature tracking unreliable. This directly increased homography estimation error.

Furthermore, the OF method became particularly unstable at extreme vehicle-to-camera distances. This vulnerability is more obvious in the VS13 scenario where vehicles move directly toward the camera, resulting in dramatic scale changes. Consequently, OF exhibits a much wider error spread in Figure \ref{fig:four_images}(b) compared to BrnoCompSpeed, where the camera is positioned overhead and the relative distance change remains minimal across the frames.

To mitigate the effects of unreliable detections at the very first and last frame, a 10\% trim was applied to each tail of the error distribution. Table \ref{tab:results_trim01} summarizes the results after this trimming. For the VS13 dataset, the keypoint method's MAE improved from 20.8\% to 14.3\%, while its within ±20\% accuracy increased from 87.6\% to 94.0\%. The optical flow method also improved from 15.0\% to 11.7\% in MAE, with within ±20\% rising from 77.9\% to 85.3\%. In terms of the BrnoCompSpeed dataset, the optical flow method's MAE dropped from 9.7\% to 7.6\%, and the keypoint method's MAE dropped from 12.8\% to 11.4\%. The trimming also significantly reduced outlier rates. For example, for the VS13 dataset, the keypoint method's outlier rate decreased from 4.4\% to 2.1\%, while the optical flow method's outlier rate decreases from 5.3\% to 2.3\%. Both method also showed enhanced performance on BrnoCompSpeed dataset after trimming.

\begin{figure}[H]
    \centering
    \small
    \begin{subfigure}[b]{0.4\linewidth}
        \centering
        \includegraphics[width=\linewidth]{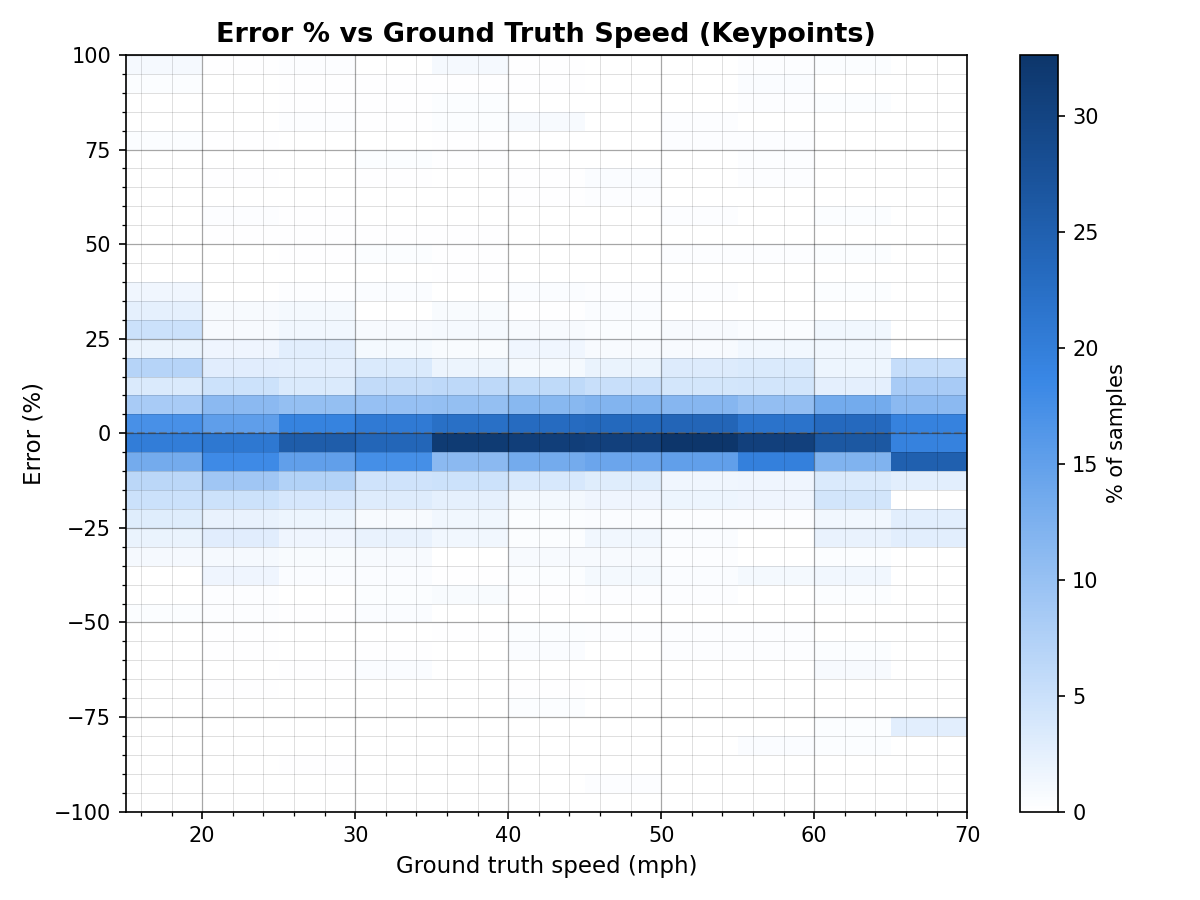}
        \caption{VS13 Keypoints}
        \label{fig:vs13_1}
    \end{subfigure}
    \hfill
    \begin{subfigure}[b]{0.4\linewidth}
        \centering
        \includegraphics[width=\linewidth]{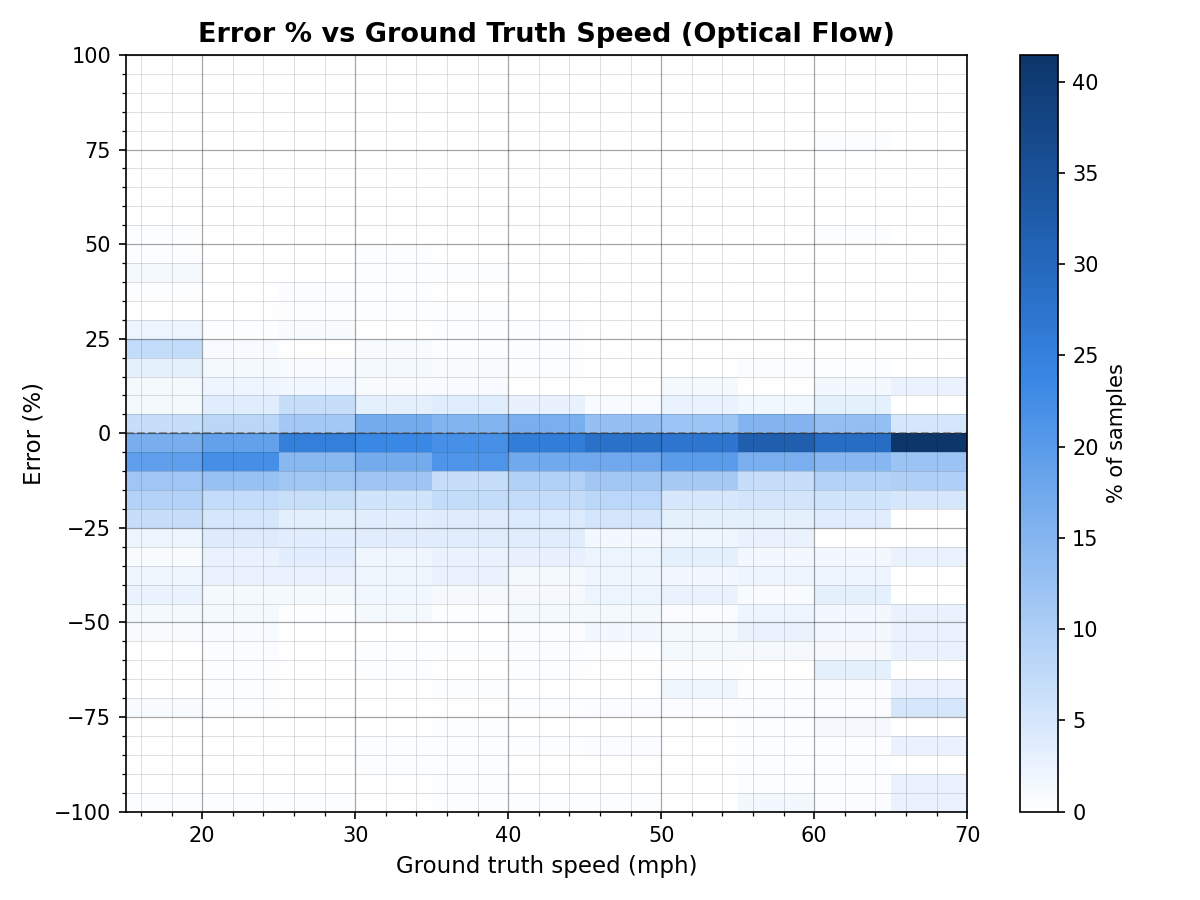}
        \caption{VS13 Optical Flow}
        \label{fig:vs13_2}
    \end{subfigure}
    \\
    \begin{subfigure}[b]{0.4\linewidth}
        \centering
        \includegraphics[width=\linewidth]{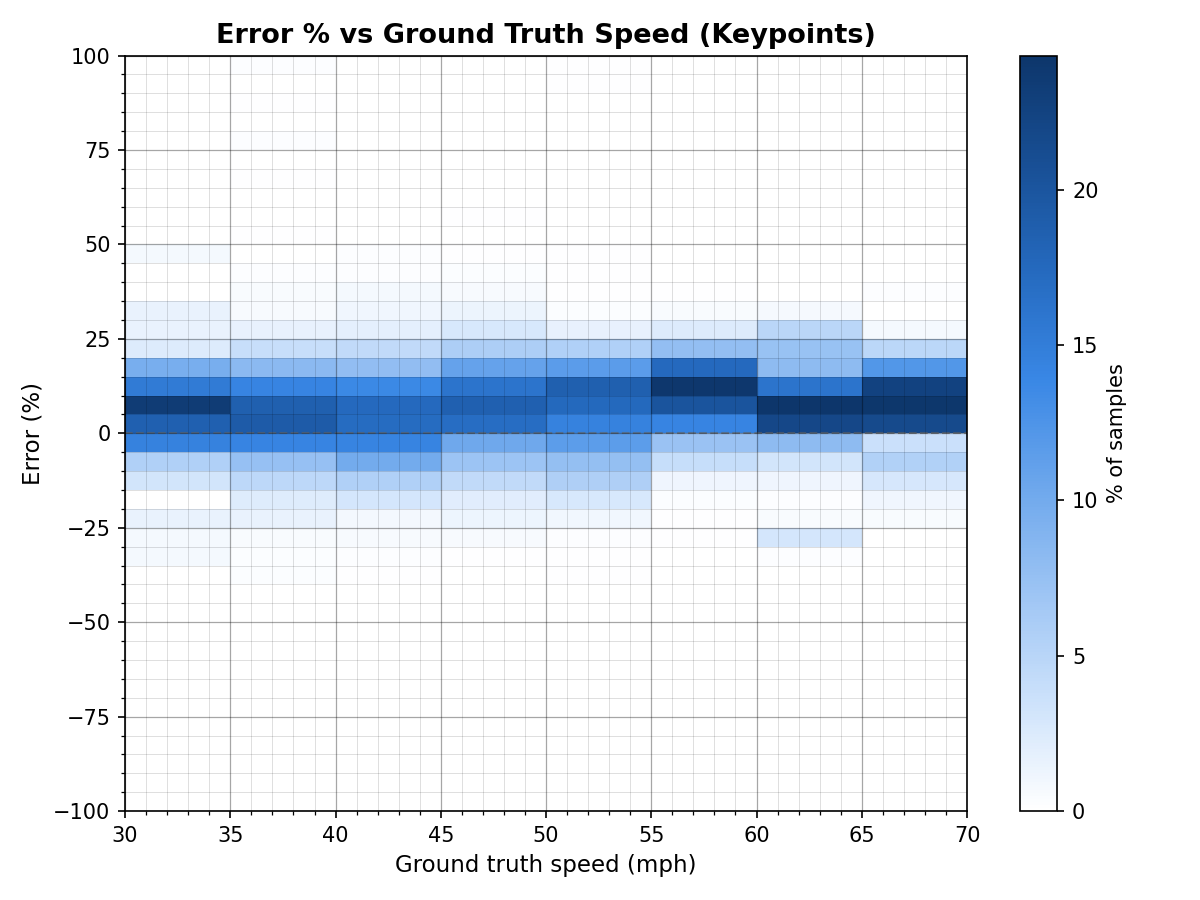}
        \caption{BrnoCompSpeed Keypoints}
        \label{fig:brno_1}
    \end{subfigure}
    \hfill
    \begin{subfigure}[b]{0.4\linewidth}
        \centering
        \includegraphics[width=\linewidth]{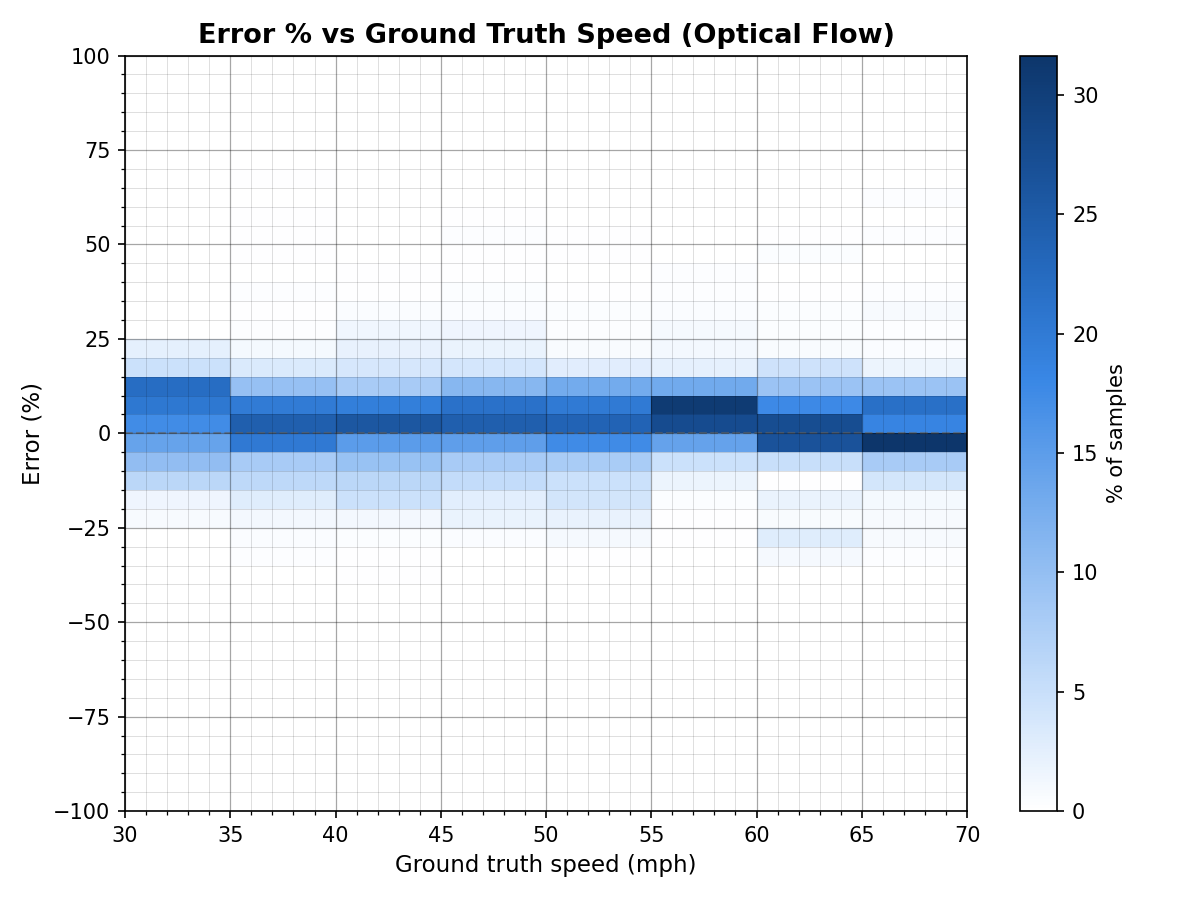}
        \caption{BrnoCompSpeed Optical Flow}
        \label{fig:brno_2}
    \end{subfigure}
    \caption{Error Percentage vs. Speed for VS13 and BrnoCompSpeed datasets}
    \label{fig:four_images_speed}
\end{figure}

Figure \ref{fig:four_images_speed} compares the speed estimation errors across different methods and datasets, with ground truth speeds ranging from 20 to 70 mph. The most accurate estimations occur in the 40 to 60 mph speed range, where both optical flow-based methods achieve errors below 5\%. In contrast, estimation errors are largest at low speeds (20 to 30 mph), particularly for keypoint-based methods, where errors exceed 20\%, possibly due to increased sensitivity to noise. 

\section{Conclusion}

This paper demonstrates that monocular camera-based vehicle speed estimation can be effectively achieved using a vehicle metric template, a lightweight vehicle keypoint detection module, and a homography technique. Crucially, the proposed method operates without requiring camera calibration (intrinsic or extrinsic) or pre-existing roadway features, relying only on a relatively stationary camera view. The approach achieves reliable and accurate speed estimation, with $87.6\%$ and $87.9\%$ of estimates falling within the $\pm 20\%$ error threshold across the evaluated datasets, demonstrating its effectiveness under diverse viewpoints and driving conditions.

\begin{table}[H]
    \centering
    \small
    \caption{Summary of speed estimation accuracy for VS13 and BrnoCompSpeed datasets with 10\% trimming.}
    \label{tab:results_trim01}
    \begin{tabular}{lllrrrrrrrr}
        \hline
        \multirow{2}{*}{\textbf{Data}} & \multirow{2}{*}{\textbf{Method}} & \multirow{2}{*}{\textbf{Group}} & \multirow{2}{*}{\textbf{N}} & \multirow{2}{*}{\textbf{Med.}} & \multirow{2}{*}{\textbf{MAE}} & \multirow{2}{*}{\textbf{RMSE}} & \multicolumn{4}{c}{\textbf{Thresholds}} \\
        \cline{8-11}
        & & & & & & & \textbf{$\pm$5\%} & \textbf{$\pm$10\%} & \textbf{$\pm$20\%} & \textbf{$>$50\%} \\
        \hline
        \multirow{6}{*}{VS13} 
            & \multirow{3}{*}{KP} 
                & All & 3,334 & -1.2\% & 14.3\% & 170.6\% & 54.0\% & 81.0\% & 94.0\% & 2.1\% \\
                & & Sedan & 1,584 & -1.2\% & 18.5\% & 240.9\% & 52.0\% & 79.9\% & 94.1\% & 1.8\% \\
                & & NoSed & 1,750 & -1.3\% & 10.5\% & 54.1\% & 55.7\% & 82.0\% & 93.9\% & 2.3\% \\
            \cline{2-11}
            & \multirow{3}{*}{OF} 
                & All & 3,555 & -5.1\% & 11.7\% & 57.4\% & 43.0\% & 66.2\% & 85.3\% & 2.3\% \\
                & & Sedan & 1,692 & -5.4\% & 11.6\% & 17.7\% & 40.1\% & 62.4\% & 81.7\% & 2.4\% \\
                & & NoSed & 1,863 & -4.8\% & 11.7\% & 77.4\% & 45.6\% & 69.7\% & 88.5\% & 2.1\% \\
        \hline
        \multirow{6}{*}{BC} 
            & \multirow{3}{*}{KP} 
                & All & 13,822 & 6.0\% & 11.4\% & 28.4\% & 28.5\% & 55.0\% & 89.4\% & 0.6\% \\
                & & Sedan & 2,738 & 6.8\% & 9.5\% & 11.9\% & 32.8\% & 60.3\% & 90.8\% & 0.0\% \\
                & & NoSed & 11,084 & 5.7\% & 11.8\% & 31.2\% & 27.4\% & 53.6\% & 89.1\% & 0.8\% \\
            \cline{2-11}
            & \multirow{3}{*}{OF} 
                & All & 14,198 & 2.3\% & 7.6\% & 11.4\% & 43.9\% & 73.9\% & 95.4\% & 0.5\% \\
                & & Sedan & 2,794 & 3.4\% & 6.5\% & 8.9\% & 50.9\% & 79.1\% & 96.7\% & 0.1\% \\
                & & NoSed & 11,404 & 1.9\% & 7.9\% & 12.0\% & 42.2\% & 72.7\% & 95.1\% & 0.6\% \\
        \hline
    \end{tabular}
    \smallskip
    {\footnotesize \textit{Note: KP = Keypoints; OF = Optical Flow. Med. = Median; MAE = Mean Absolute Error; RMSE = Root Mean Square Error; N = Number of frames with valid estimation.}}
\end{table}

Experimental results also indicate that correspondence computation via the proposed warped optical flow method is more reliable and robust against outliers. In contrast, purely keypoint-based correspondence can offer high peak accuracy but remains sensitive to keypoint detection jitter.

Additionally, this work has significant practical implications for traffic enforcement and public safety. Unlike traditional speed measurement systems that rely on roadway features, the proposed method is deployable and cost-effective. This facilitates citizen-based enforcement, where the public can use portable devices, such as dashcams or smartphones, to document speeding events \cite{li2025smartphone}. Nowadays, many cities have piloted citizen-based enforcement programs. For example, New York City has empowered citizens to enforce diesel-truck idling laws by submitting video evidence; those who provide at least three minutes of footage receive 25\% of any fine collected, amounting to about \$87.50 per violation \cite{hollis2026trucking}. As vision-based enforcement becomes increasingly popular due to budget constraints and the rapid growth of camera-equipped devices, the proposed approach delivers a solution that can complement existing enforcement mechanisms while reducing the cost of camera installation.

However, the current implementation has certain limitations. It relies on a generalized sedan metric template, which can introduce minor errors due to keypoint detection inaccuracies and homography transformations. Although experiments show that the framework maintains decent accuracy on non-sedan vehicles, future work will focus on expanding the template library to include diverse vehicle classes and geometries. Additionally, while the warped optical flow approach mitigates individual outliers, achieving even higher accuracy will likely require more robust and reliable correspondence tracking methods to handle challenging conditions such as illumination changes or vehicles in close proximity to the camera.

\section{Acknowledgments}
The authors thank Helena Chandy for her valuable insights and edits.

% \section*{AUTHOR CONTRIBUTIONS}
% The authors confirm contribution to the paper as follows: 
% study conception and design: X.~Author, Y.~Author; 
% data collection: Y.~Author; 
% analysis and interpretation of results: X.~Author, Y.~Author, Z.~Author; 
% draft manuscript preparation: Y.~Author, Z.~Author. 
% All authors reviewed the results and approved the final version of the manuscript.

\section*{DECLARATION OF CONFLICTING INTERESTS}
All authors declare no potential conflicts of interest with respect to the research, authorship, and publication of this article.

\newpage
\bibliographystyle{trb}
\bibliography{trb_template}
\end{document}